\documentclass[journal]{IEEEtran}
\usepackage{amsmath,amsfonts}
\usepackage{algorithmic}
\usepackage{algorithm}
\usepackage{array}
\usepackage{textcomp}
\usepackage{stfloats}
\usepackage{url}
\usepackage{verbatim}
\usepackage{graphicx}
\usepackage{epstopdf}
\usepackage{cite}
\usepackage{booktabs}
\usepackage{adjustbox}
\usepackage{multirow}
\usepackage{hyperref}
\hypersetup{hidelinks}
\usepackage{mathtools}
\usepackage{adjustbox}
\usepackage{multirow}
\usepackage{enumitem}
\usepackage{multirow}
\usepackage{diagbox}
\usepackage{bm}
\ifCLASSINFOpdf
\else
   \usepackage[dvips]{graphicx}
\fi
\usepackage{url}

\usepackage{graphicx}
\usepackage{hyperref}

\begin{document}

\title{Video Inpainting Localization with Contrastive Learning}

\author{Zijie Lou, Gang Cao, \IEEEmembership{Member, IEEE}, and Man Lin 
\thanks{Zijie Lou, Gang Cao and Man Lin are with the State Key Laboratory of Media Convergence and Communication, Communication University of China, Beijing 100024, China, and also with the School of Computer and Cyber Sciences, Communication University of China, Beijing 100024, China (e-mail: \{louzijie2022, gangcao, linman\}@cuc.edu.cn).}}

\markboth{Journal of \LaTeX\ Class Files, Vol. 14, No. 8, August 2015}
{Shell \MakeLowercase{\textit{et al.}}: Bare Demo of IEEEtran.cls for IEEE Journals}
\maketitle

\begin{abstract}
Deep video inpainting is typically used as malicious manipulation to remove important objects for creating fake videos. It is significant to identify the inpainted regions blindly. This letter proposes a simple yet effective forensic scheme for Video Inpainting LOcalization with ContrAstive Learning (ViLocal). Specifically, a 3D Uniformer encoder is applied to the video noise residual for learning effective spatiotemporal forensic features. To enhance the discriminative power, supervised contrastive learning is adopted to capture the local inconsistency of inpainted videos through attracting/repelling the positive/negative pristine and forged pixel pairs. A pixel-wise inpainting localization map is yielded by a lightweight convolution decoder with a specialized two-stage training strategy. To prepare enough training samples, we build a video object segmentation dataset of 2500 videos with pixel-level annotations per frame. Extensive experimental results validate the superiority of ViLocal over state-of-the-arts. Code and dataset will be available at \href{https://github.com/multimediaFor/ViLocal}{https://github.com/multimediaFor/ViLocal}.
\end{abstract}

\begin{IEEEkeywords}
Video Forensics, Video Inpainting Localization, Contrastive Learning, Video Transformer
\end{IEEEkeywords}

\IEEEpeerreviewmaketitle

\section{Introduction}

\label{sec:intro}
\setlength{\parindent}{1em}
Video inpainting aims to repair missing or damaged regions with plausible and coherent contents in a video. 
However, video inpainting may also be used to create forged videos by deleting or altering some contents, such as falsifying forensic evidences, removing copyright marks or key objects in news videos. Such malicious use of video inpainting potentially incurs societal risks and legal concerns. 
Therefore, it is necessary to develop reliable forensic methods to identify the inpainted regions in a video for realizing refined authentication.  


Actually, one of the most important things in video inpainting localization is to discover the spatiotemporal inconsistency between patches and frames. Yang \textit{et al.} \cite{yang2020spatiotemporal} explore local spatiotemporal features by applying 3D convolution, which can reduce spatiotemporal redundancy across adjacent frames. However, such 3D convolution suffers from difficulty in learning long-range dependency due to the limited receptive field \cite{wang2018non}. Yu \textit{et al.} \cite{yu2021frequency} capture the spatial artifacts of inpainted frames by vision transformer, which is good at learning global dependency with the help of self-attention mechanism. Recently, Li \textit{et al.} \cite{li2023uniformer} point out that naive transformer can learn detailed video representations, but with very redundant spatial and temporal attention. In addition, \cite{zhou2021deep} and \cite{wei2022deep} employ traditional convolution and LSTM (Long Short Term Memory) to model spatiotemporal relationships, where the direct and hard combination may lead to inconsistent predictions. 

Another major disadvantage of existing video inpainting localization methods \cite{zhou2021deep, yu2021frequency, wei2022deep, nguyen2022videofact, yao2024deep, pei2023uvl, pei2023vifst} relies on adopting the cross-entropy (CE) loss without additional constraints for training. As pointed out by \cite{zhao2020makes}, the traditional CE-based methods assume that all instances within each category should be close in feature distribution. However, extracting similar features for all the inpainted regions in video datasets is infeasible as different inpainting algorithms may leave behind distinct forgery traces. Hence, the CE-based framework without reasonable constraints is prone to over-fitting on some specific inpainting algorithms. This is not conducive to generalization.

\begin{figure}[!t]
\centering
\includegraphics[width=\linewidth]{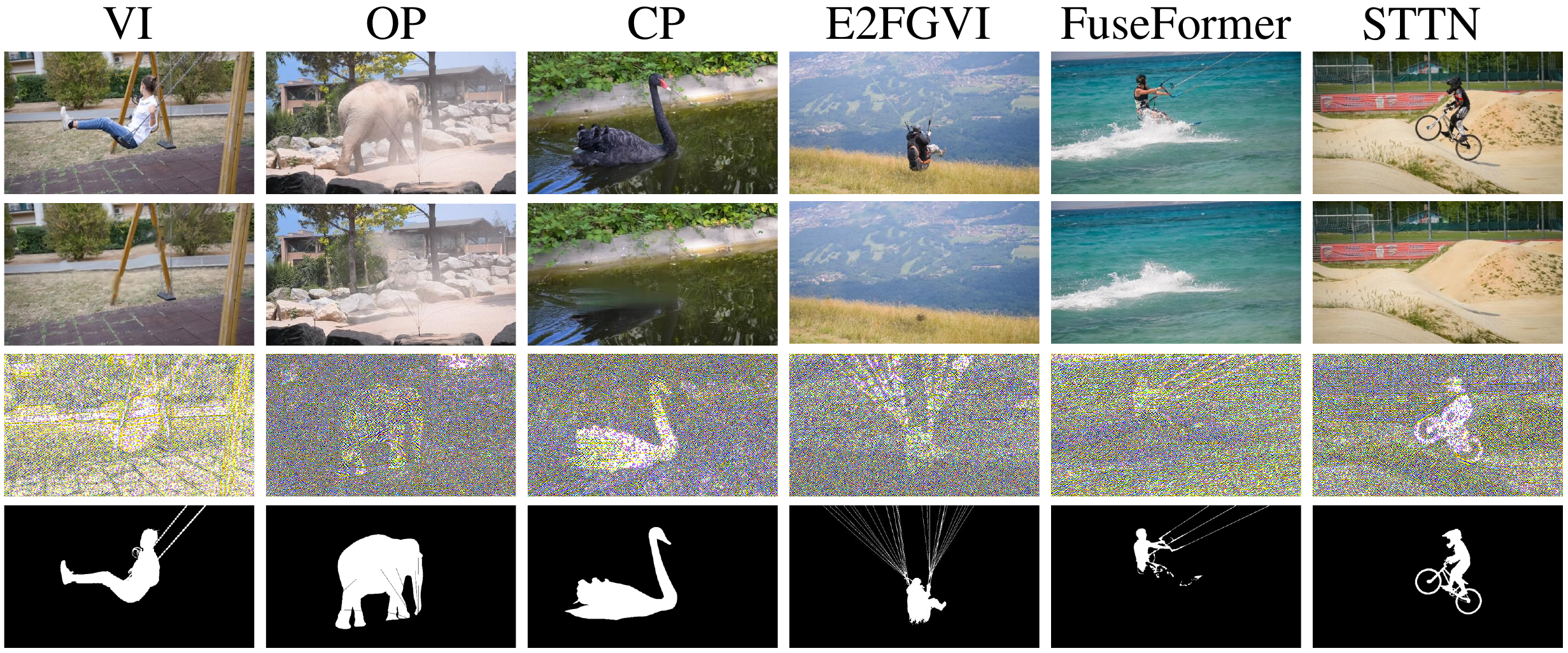}
\caption{Illustration of inpainting artifacts extracted by HP3D. From top to bottom: original frames, inpainted frames and corresponding noise images, and ground truth. 
}
\label{noise_feature}
\vspace{-0.3cm}
\end{figure}

To attenuate the deficiencies of prior works, here we propose a simple yet effective forensic framework based on video transformer and contrastive learning. Specifically, 3D Uniformer \cite{li2023uniformer} is modified as the backbone encoder network, which integrates merits of 3D convolution and spatiotemporal self-attention. A lightweight convolution decoder is designed to generate the pixel-wise inpainting localization map. As shown in Fig. \ref{noise_feature}, the noise domain yielded by HP3D filter \cite{lou2024trusted} is explored to expose the inpainting traces distinctly. An effective two-stage protocol is adopted to train the encoder and decoder networks separately. Extensive evaluations verify the robustness and generalization ability of our proposed ViLocal scheme. 

The rest of this letter is organized as follows. The proposed video inpainting localization method is described in Section II, followed by extensive experimental results and discussions in Section III. We draw the conclusions in Section IV.



\section{Proposed ViLocal Scheme}
\label{sec:method}

\begin{figure*}[!t]
\centering
\includegraphics[width=\textwidth]{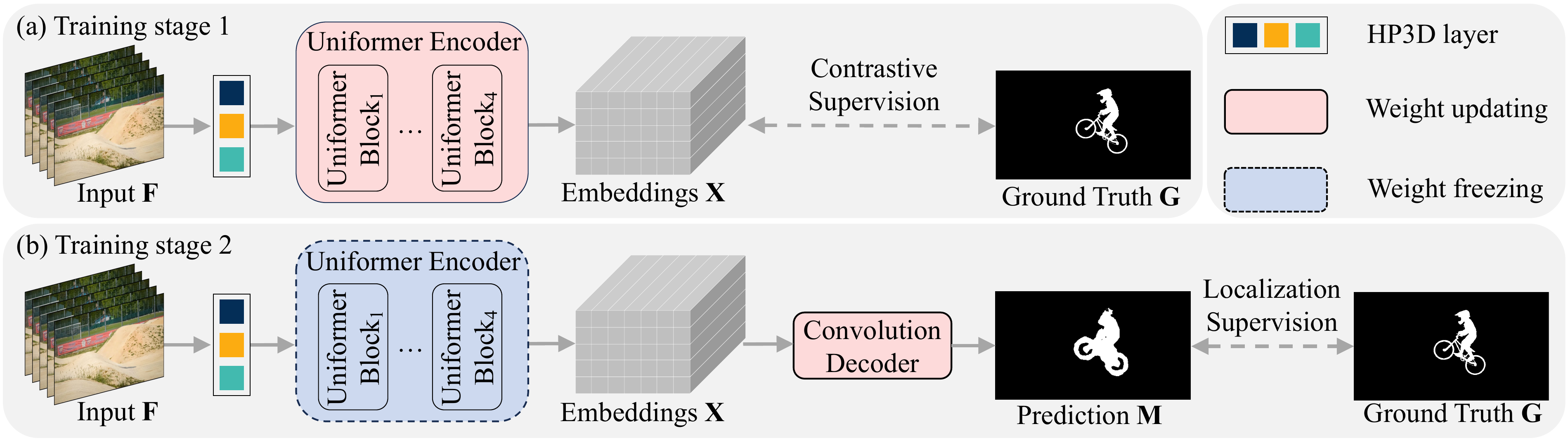}
\caption{Proposed video inpainting localization scheme ViLocal. Each 5 consecutive frames is set as an input unit to yield the inpainting localization map of the middle frame. (a) Training stage 1. ViLocal utilizes contrastive supervision to train the encoder network. (b) Training stage 2. ViLocal employs localization supervision to train the decoder network. }
\label{framework}
\end{figure*}

\subsection{Network Architecture}
The network architecture of ViLocal is shown in Fig. \ref{framework}.
Given an inpainted video sequence $\textbf{F}=\{f_{1}, f_{2}, \dotsb, f_{T}\}$ with $\mathit{T}$ consecutive frames, where $f_i \in \mathbb{R}^{H \times W \times C}$. The Uniformer encoder \cite{li2023uniformer} takes its HP3D noise feature as the input, and outputs the embeddings $\textbf{X} \in \mathbb{R}^{\frac{H}{4} \times \frac{W}{4} \times 256}$. To limit the computational cost of whole localization network, we design a lightweight convolution decoder. The embeddings $\mathrm{\mathbf{X}}$ extracted by the 3D Uniformer encoder is first mapped to $\mathbb{R}^{\frac{H}{4} \times \frac{W}{4} \times 128}$ by a $ 1 \times 1 $ convolution layer. A batch normalization (BN) layer and a rectified linear unit (ReLU) layer are configured subsequently. Then, another $ 1 \times 1 $ convolution layer followed by a bilinear upsampling layer and a sigmoid layer is deployed to generate the predicted localization probability map $\mathrm{\mathbf{M}} \in [0, 1]^{H \times W}$. Finally, the threshold is applied to $\mathrm{\mathbf{M}}$ for getting the binary localization result. 

Specifically, for each input data unit, noise features are extracted by HP3D layer \cite{lou2024trusted} firstly. In the first training stage, Uniformer encoder is employed to extract spatiotemporal features from noise view. Then the relationships of local features between inpainted and authentic regions are exploited for discriminative representation learning through contrastive loss. In the second training stage, weights of the trained encoder network are frozen. The decoder network is optimized guided by the focal loss \cite{lin2017focal} which can diminish the effect of class imbalance existing in the datasets. 

\begin{figure}[!t]
\centering
\includegraphics[width=\linewidth]{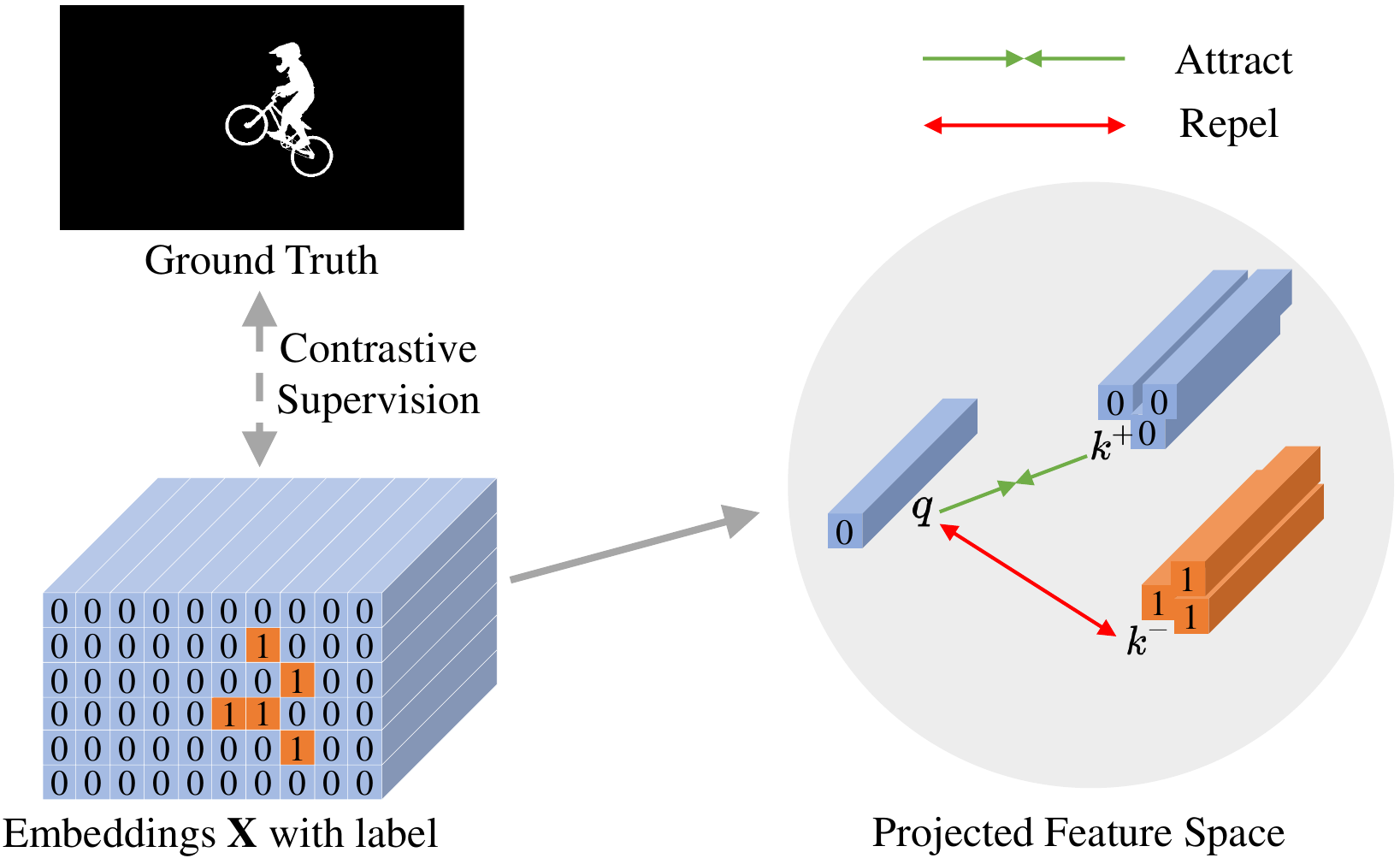}
\caption{Illustration of our supervised contrastive learning for video inpainting localization.}
\label{CL}
\vspace{-0.6cm}
\end{figure}

\subsection{Two-stage Training Protocol}
\noindent 
\textbf{Contrastive Supervision.}
We aim to contrast between the inpainted and authentic pixel embeddings of each sample. For a pixel embedding $ q \in \mathrm{\mathbf{X}}$ with groundtruth label "0", the positive samples $k^+$ are the other pixel embeddings with label "0" in $\mathrm{\mathbf{X}}$, while the negative samples $k^-$ are the ones with label "1" in $\mathrm{\mathbf{X}}$. The supervised contrastive loss is defined as

\begin{align}
\mathcal{L}_{\mathrm{Contra.}}= -_{\!}\log\frac{\frac{1}{|\mathcal{P}_k|} \sum_{k^+ \in \mathcal{P}_k} \exp(q \cdot k^{+}/\tau)}{\sum\nolimits_{k^{-}\in\mathcal{N}_k} \exp(q \cdot k^{-}/\tau)} \tag{1}
\end{align}

\noindent
where $\mathcal{P}_x$ and $\mathcal{N}_x$ denote the positive and negative pixel embedding collections randomly sampled from $\mathrm{\mathbf{X}}$, respectively. ‘·’ denotes the inner product, and $\tau$ is a temperature hyper-parameter. Such contrastive loss is aimed at learning discriminative feature representation, which benefits to distinguish the pristine and forged pixels. By pulling the same class of pixel embeddings close and pushing different class of pixel embeddings apart, the intra-class compactness and inter-class separability can be improved \cite{niloy2023cfl, zhou2023pre}.

\noindent 
\textbf{Localization Supervision.}
Once the encoder network is trained, its weights are frozen in the latter training. In this stage, the goal is to map the encoded forensic representation $\mathrm{\mathbf{X}}$ to a pixel-wise inpainting localization mask. 
Specifically, the convolution decoder is optimized by the focal loss \cite{lin2017focal}:
\begin{align}
    \begin{split}
        &\mathcal{L}_{\mathrm{Focal}}(\mathrm{\mathbf{G}},\mathrm{\mathbf{M}})= \begin{aligned}-\sum\alpha\left(1-\mathrm{\mathbf{M}}_{ij}\right)^{\gamma}*\mathrm{\mathbf{G}}_{ij}\log\left(\mathrm{\mathbf{M}}_{ij}\right)\end{aligned}  \\
        &\begin{aligned}-\sum(1-\alpha)\mathrm{\mathbf{M}}_{ij}^{\gamma}*(1-\mathrm{\mathbf{G}}_{ij})\log{(1-\mathrm{\mathbf{M}}_{ij})},   \end{aligned}
    \end{split} \tag{2}
\end{align} 

\noindent 
where $\mathrm{\mathbf{G}}_{ij} \in \{0, 1\}$ denotes the pixel-level label and $\mathrm{\mathbf{M}}_{ij} \in [0, 1]$ denotes the corresponding prediction result. $\alpha$ and $\gamma$ are the hyperparameters.






\section{Experiments}
\label{sec:experiments}

\begin{table*}[!t]
\centering
\tabcolsep=10 pt
\caption{Datasets used in our experiments. We indicate the inpainting datasets and their source datasets.}
\begin{adjustbox}{width=\textwidth}
\begin{tabular}{rccc|ccc}
\toprule[1pt]
\textbf{Source Dataset}     &\textbf{Videos}     & \textbf{Frames}   & \textbf{Applied Inpainting Algorithms}    & \textbf{Inpainting Dataset}  & \textbf{Videos}  & \textbf{Usage}        \\    
\midrule
VOS2k5                   & 800          & 102843                & VI, OP, CP        & \multirow{3}{*}{DVI2016} & \multirow{2}{*}{(800$+$30)$\times$3} & \multirow{2}{*}{train} \\
\multirow{2}{*}{DAVIS2016 \cite{perazzi2016benchmark}} & 30    & 2079     & VI, OP, CP         &           &                 &                        \\
                                                      & 20    & 1376     & VI, OP, CP         &           & 20$\times$3              & test          \\ 
\midrule
\multirow{2}{*}{DAVIS2017 \cite{Pont-Tuset_arXiv_2017}} & \multirow{2}{*}{90} & \multirow{2}{*}{6208} & \multirow{2}{*}{E2FGVI, FuseFormer, STTN}  & \multirow{2}{*}{DVI2017} & \multirow{2}{*}{90$\times$3}       & \multirow{2}{*}{test}  \\
                           &                     &                       &                                                        &                          &                             &                        \\
MOSE \cite{MOSE}                       & 100                 & 8183                  & E2FGVI, FuseFormer, STTN                               & MOSE100                  & 100$\times$3                      & test                  \\
\bottomrule[1pt]
\end{tabular}
\end{adjustbox}
\label{tab-dataset}
\end{table*}

\begin{table*}[!t]
\centering
\caption{Accuracy comparison of different inpainting localization methods on DVI2016 dataset. All methods are trained on the datasets inpainted by VI and OP, OP and CP, VI and CP algorithms, respectively (denoted as ‘*’). ‘-’ denotes that the result is not available. Bold numbers represent the best results.}
\begin{adjustbox}{width=\textwidth}
\begin{tabular}{rccccccccc}
\toprule[1pt]
              & \textbf{VI*}       & \textbf{OP*}       & \textbf{CP}                             & \textbf{VI}        & \textbf{OP*}       & \textbf{CP*}                            & \textbf{VI*}       & \textbf{OP}        & \textbf{CP*}       \\
Methods       & IoU/F1             & IoU/F1             & \multicolumn{1}{c|}{IoU/F1}             & IoU/F1             & IoU/F1             & \multicolumn{1}{c|}{IoU/F1}             & IoU/F1             & IoU/F1             & IoU/F1             \\ 
\midrule
NOI{ \cite{mahdian2009using}}     & 0.08/0.14          & 0.09/0.14          & \multicolumn{1}{c|}{0.07/0.13}          & 0.08/0.14          & 0.09/0.14          & \multicolumn{1}{c|}{0.07/0.13}          & 0.08/0.14          & 0.09/0.14          & 0.07/0.13          \\
CFA{ \cite{ferrara2012image}}     & 0.10/0.14          & 0.08/0.14          & \multicolumn{1}{c|}{0.08/0.12}          & 0.10/0.14          & 0.08/0.14          & \multicolumn{1}{c|}{0.08/0.12}          & 0.10/0.14          & 0.08/0.14          & 0.08/0.12          \\
COSNet{ \cite{lu2019see}}  & 0.40/0.48          & 0.31/0.38          & \multicolumn{1}{c|}{0.36/0.45}          & 0.28/0.37          & 0.27/0.35          & \multicolumn{1}{c|}{0.38/0.46}          & 0.46/0.55          & 0.14/0.26          & 0.44/0.53          \\
HP-FCN{ \cite{li2019localization}}     & 0.46/0.57          & 0.49/0.62          & \multicolumn{1}{c|}{0.46/0.58}          & 0.34/0.44          & 0.41/0.51          & \multicolumn{1}{c|}{0.68/0.77}          & 0.55/0.67          & 0.19/0.29          & 0.69/0.80          \\
GSR-Net{ \cite{zhou2020generate}} & 0.57/0.69          & 0.50/0.63          & \multicolumn{1}{c|}{0.51/0.63}          & 0.30/0.43          & 0.74/0.82          & \multicolumn{1}{c|}{0.80/0.85}          & 0.59/0.70          & 0.22/0.33          & 0.70/0.77          \\
VIDNet{ \cite{zhou2021deep}}  & 0.59/0.70          & 0.59/0.71          & \multicolumn{1}{c|}{0.57/0.69}          & 0.39/0.49          & 0.74/0.82          & \multicolumn{1}{c|}{0.81/0.87}          & 0.59/0.71          & 0.25/0.34          & 0.76/0.85          \\
USTT{ \cite{wei2022deep}}  & 0.60/0.73 & 0.69/0.80          & \multicolumn{1}{c|}{0.65/0.77}          & -                  & -                  & \multicolumn{1}{c|}{-}                  & -                  & -                  & -                  \\
FAST{ \cite{yu2021frequency}}    & 0.61/0.73 & 0.65/0.78          & \multicolumn{1}{c|}{0.63/0.76}          & 0.32/0.49          & 0.78/0.87          & \multicolumn{1}{c|}{0.82/\textbf{0.90}} & 0.57/0.68          & 0.22/0.34          & 0.76/0.83          \\
UVL{ \cite{pei2023uvl}}    & \,\,0.65/\,\; - \;\; & \,\,0.66/\;\; - \;\;          & \multicolumn{1}{c|}{\,\,0.65/\;\; - \;\;}     & \textbf{\,\,0.64}/\;\; - \;\;  & \,\,0.67/\;\; - \;\;  & \multicolumn{1}{c|}{\,\,0.68/\;\; - \;\;} & \textbf{\,\,0.75}/\;\; - \;\;          & \textbf{\,\,0.75}/\;\; - \;\;          & \,\,0.74/\;\; - \;\;          \\
ViLocal (Ours)          & \textbf{0.66/0.77} & \textbf{0.82/0.89} & \multicolumn{1}{c|}{\textbf{0.76/0.85}} & 0.61/\textbf{0.73} & \textbf{0.85/0.91} & \multicolumn{1}{c|}{\textbf{0.83/0.90}} & 0.69/\textbf{0.79} & 0.63/\textbf{0.75} & \textbf{0.82/0.89} \\ 
\bottomrule[1pt]
\end{tabular}
\end{adjustbox}
\label{Results_DVI2016}
\end{table*}



In this section, extensive experiments are performed to verify the effectiveness of ViLocal. The following datasets are used in our experiments, details are shown in Table \ref{tab-dataset}. DAVIS2016 \cite{perazzi2016benchmark} and DAVIS2017 \cite{Pont-Tuset_arXiv_2017} are the most famous benchmarks for deep video inpainting. In addition, we create VOS2k5 dataset from YouTube-VOS \cite{Xu_2018_ECCV} and GOT10k \cite{8922619}, and create MOSE100 dataset from MOSE \cite{MOSE}.

ViLocal is implemented using the PyTorch deep learning framework. Each 5 consecutive frames is set as an input unit and the batch size is 2. All the frames used in training are resized to $432 \times 240$ pixels. And one quarter of the training set is compressed by H.264 with the Constant Rate Factor (CRF) 23. We train the network on an A800 GPU for two stages. F1 and IoU are used as the metrics of pixel-level localization accuracy. For calculating F1 and IoU, the threshold is set to 0.5 by default.

\subsection{Compared with State-of-the-Art Methods}
We first compare the performance of ViLocal with several related methods on DVI2016 dataset. Two versions of DVI2016 datasets are used for training and in-domain testing. And the other one is used for cross-domain testing. 
Table \ref{Results_DVI2016} shows the quantitative comparison results 
For all the three training settings, ViLocal outperforms other approaches on all trained video inpainting approaches. It presents the advantages of our approach to acquire inpainting artifacts distributed in the videos. Furthermore, ViLocal also exceeds other approaches on all the unseen video inpainting approaches. It presents the powerful generalization ability of our approach. For example, when trained on VI and CP methods, the IoU and F1 on OP associated with VIDNet and FAST only reach about $0.2$ and $0.3$. In contrast, our proposed method still has high accuracy, its IoU and F1 reach 0.63 and 0.75, respectively. Compared to UVL \cite{pei2023uvl}, our ViLocal continues to maintain a leading advantage in both in-domain and cross-domain testing.

\begin{figure}
\centering
\includegraphics[width=\linewidth]{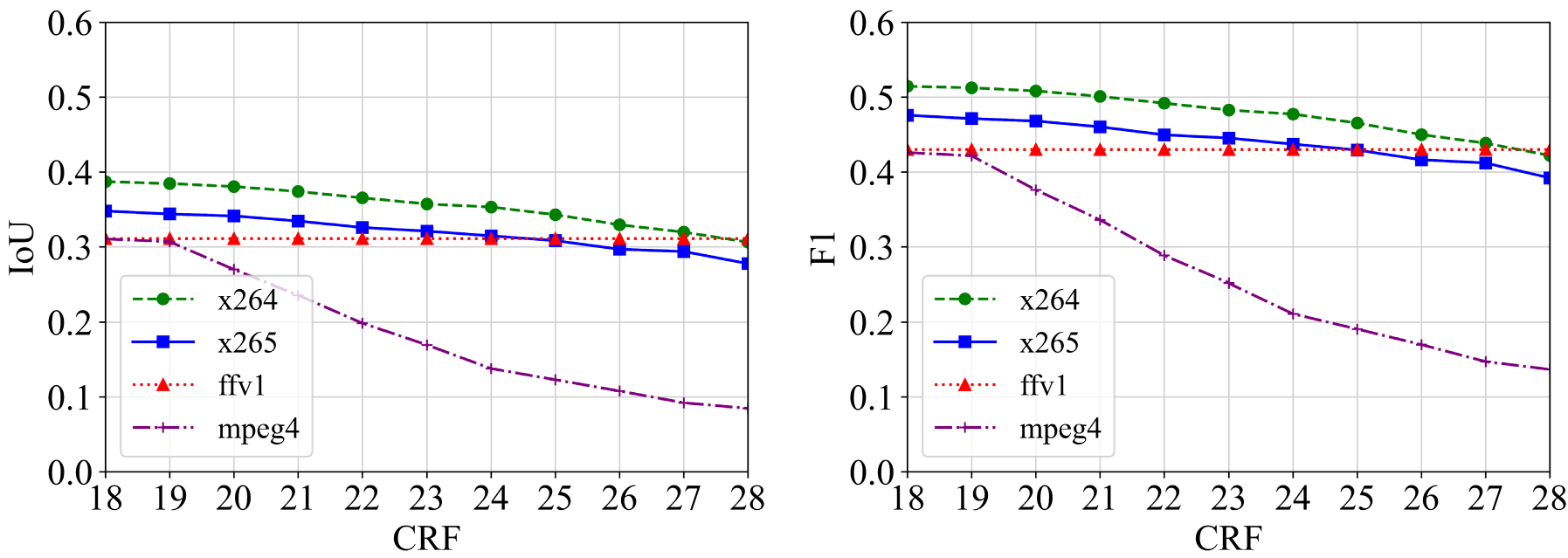}
\caption{IoU and F1 for different codecs and CRFs.}
\label{diff_codec}
\end{figure}

\begin{table}[!t]
\centering
\caption{IoU and F1 for video recompression. The CRF is fixed at 23.}
\begin{adjustbox}{width=\linewidth}
\begin{tabular}{l|cccc}
\toprule
\diagbox [width=6em,trim=l] {2nd com.}{1st com.} & x264 & x265 & ffv1  & mpeg4 \\
\hline
x264  & 0.341/0.463  & 0.331/0.451 & 0.362/0.488 & 0.282/0.396   \\
x265  & 0.303/0.422  & 0.308/0.429 & 0.321/0.445 & 0.266/0.379  \\
ffv1  & 0.343/0.465  & 0.309/0.429 & 0.307/0.424 & 0.152/0.227   \\
mpeg4 & 0.155/0.235  & 0.146/0.223 & 0.167/0.248 & 0.163/0.244   \\
\bottomrule
\end{tabular}\vspace{0cm}
\end{adjustbox}
\label{recompression}
\end{table}

\begin{figure*}[!t]
\centering
\includegraphics[width=\textwidth]{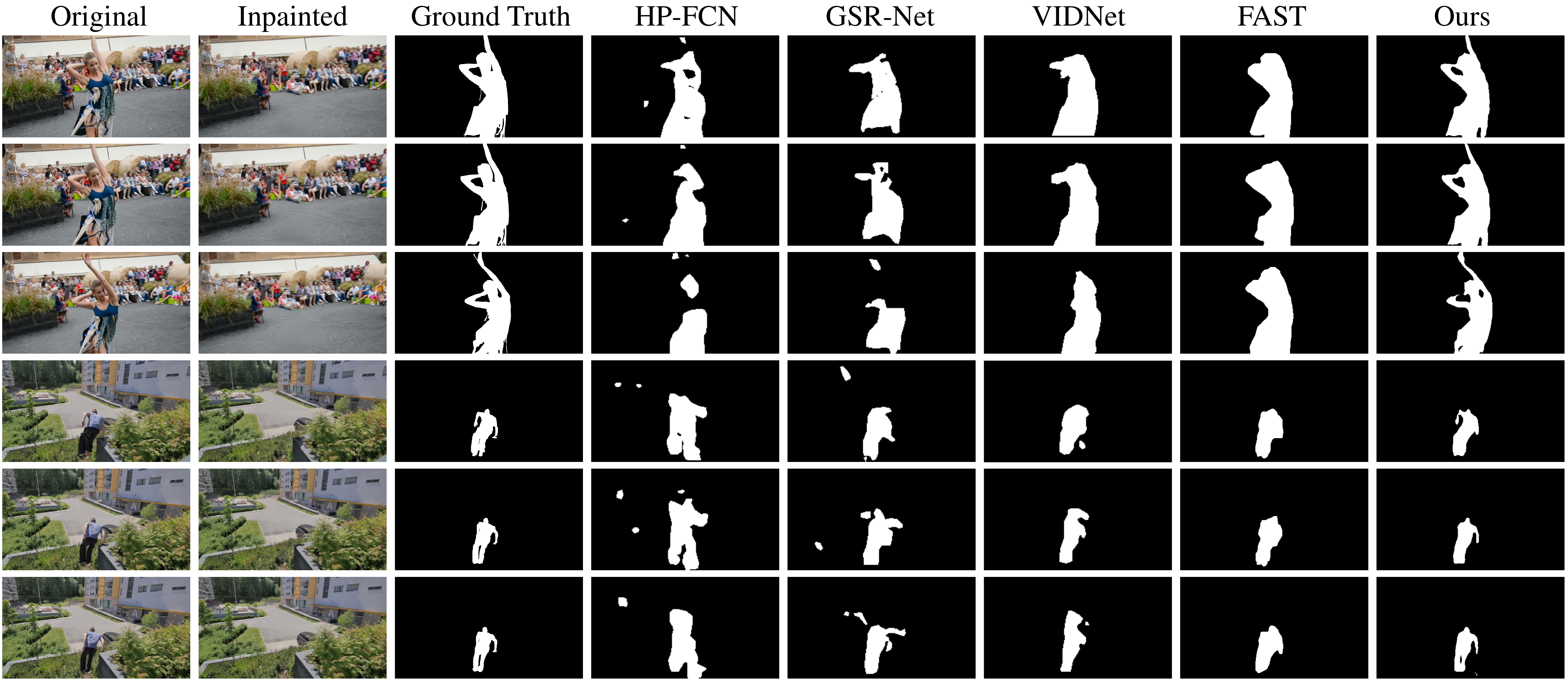}
\caption{Qualitative visualization on two DVI2016 videos.}
\label{fig_5}
\end{figure*}

\subsection{Robustness Evaluation}
To simulate video forensics in real-world scenarios, inpainted videos are compressed by four of the most popular codecs built into FFmpeg. Our model is trained on DVI2016 dataset inpainted by VI and OP algorithms. Fig. \ref{diff_codec} shows the performance of ViLocal under different codecs compression. It can be observed that ViLocal exhibits certain degrees of robustness against compression. Further, we compress the inpainted videos twice to test the robustness of our model against video recompression. As shown in Table \ref{recompression}, our model can resist a certain degree of double compression. 


\begin{table}[!t]
\centering
\caption{Accuracy comparison of different inpainting localization methods on DVI2017 and MOSE100 datasets. Our model is trained on DVI2016 dataset inpainted by VI and OP algorithms.}
\tabcolsep=8pt
\begin{adjustbox}{width=\linewidth}
\begin{tabular}{rcccccc}
\toprule[1pt]
                                           & \textbf{E2FGVI} {\cite{li2022towards}}        & \textbf{FuseFormer} {\cite{liu2021fuseformer}}    & \textbf{STTN} {\cite{zeng2020learning}}   \\
Datasets                                    & IoU/F1               & IoU/F1               & IoU/F1               \\
\midrule
DVI2017                        & 0.402/0.528      & 0.592/0.712       & 0.387/0.515 \\
Mose100                             & 0.485/0.620 & 0.597/0.721 & 0.393/0.524 \\
\bottomrule[1pt]
\end{tabular}
\end{adjustbox}

\label{general}
\end{table}

\begin{table}[!t]
\centering
\caption{Evaluation of different components of ViLocal. The models are trained on DVI2016 inpainted by VI and OP algorithms (denoted as ‘*’) and tested on DVI2016 dataset inpainted by CP.}
\tabcolsep=9 pt
\begin{adjustbox}{width=\linewidth}
\begin{tabular}{lccc}
\toprule[1pt]
                                           & \textbf{VI*}         & \textbf{OP*}         & \textbf{CP}  \\
Methods                                    & IoU/F1            & IoU/F1            & IoU/F1    \\
\midrule
Ours w/o HP3D                               & 0.62/0.71          & 0.75/0.82          & 0.68/0.79   \\
Ours w/o $\mathcal{L}_{\mathrm{Contra}}$   & 0.45/0.54          & 0.58/0.66          & 0.54/0.63          \\
ViLocal (Ours)                             & \textbf{0.66/0.77} & \textbf{0.82/0.89} & \textbf{0.76/0.85} \\
\bottomrule[1pt]
\end{tabular}
\end{adjustbox}
\label{Ablation}
\end{table}

\subsection{Generalization Analysis}
In this section, we extensively test the generalization performance of the proposed method. Due to the unavailability of code and pre-trained models for most video inpainting localization algorithms, we evaluated our method only under two experimental settings: cross-inpainting algorithm (on DVI2017 dataset) and cross-dataset (on Mose100 dataset). As shown in Table \ref{general}, although there are significant differences in various inpainting algorithms and datasets, ViLocal exhibits excellent generalization ability thanks to contrastive learning.

\subsection{Ablation Studies}
We conduct extensive ablation studies to analyze how each component of our model contributes to the final localization results. 
First, the HP3D layer is dropped to demonstrate the benefit of noise features. We can observe that the performance of our model deteriorates without HP3D layer. This is mainly because noise features can unveil the inpainting traces, providing powerful evidence for inpainting localization. Then, 
we train the entire video inpainting localization network solely based on the localization loss. The results show that our proposed ViLocal without contrastive learning perform worse. It illustrates the importance of contrastive learning, which can focus on the essential inpainting-caused feature differences.

\subsection{Qualitative Results}
Fig. \ref{fig_5} illustrates the visualization of our predictions compared with others under the same setting. It is observed that ViLocal predicts the masks that are closest to the ground truth. Specifically, 
Though VIDNet achieves temporally consistent predictions by convolutional LSTM, its localization results exhibit occasional omissions of intricate details. Based on the extraction of frequency-aware features, FAST greatly improves the localization performance in spatial details, but it is prone to false alarms. Compared with these methods, our proposed ViLocal generates more precise predicted masks.

\section{Conclusion}
\label{sec:conclusion}

In this paper, we 
propose a simple yet effective video inpainting localization network, which consists of a 3D Uniformer encoder and a lightweight convolution decoder. The encoder and decoder networks are trained with contrastive and localization losses, respectively. We have performed extensive performance evaluations on various datasets, inpainting algorithms, and post-processing. The results have demonstrated the high localization accuracy, impressive generalization ability, and robustness of our proposed ViLocal. In the future work, it is expected to expand such a video inpainting forensic scheme to deal with more types of video forgeries, such as video splicing and copy-move. 
\bibliographystyle{IEEEtran}
\bibliography{reference}

\end{document}